\documentclass{ieeetj}
\usepackage{cite}
\usepackage{amsmath,amssymb,amsfonts}
\usepackage{algorithmic}
\usepackage{graphicx,color}
\usepackage{textcomp}
\usepackage{xcolor}
\usepackage{hyperref}
\hypersetup{hidelinks=true}
\usepackage{algorithm,algorithmic}
\def\BibTeX{{\rm B\kern-.05em{\sc i\kern-.025em b}\kern-.08em
    T\kern-.1667em\lower.7ex\hbox{E}\kern-.125emX}}
\AtBeginDocument{\definecolor{tmlcncolor}{cmyk}{0.93,0.59,0.15,0.02}\definecolor{NavyBlue}{RGB}{0,86,125}}

\def\OJlogo{}
\def\seclogo{}

\def\authorrefmark#1{\ensuremath{^{\textbf{#1}}}}

\begin{document}

\markboth{}{Author {et al.}}

\title{PTQ-NAS-Bench: A Multi-Scheme Post-Training Quantization Benchmark for Hardware-Efficient Architecture and Scheme Co-Search}

\author{Eleftherios Mylonas\authorrefmark{1}, Alexios Birbas\authorrefmark{1}, Member, IEEE, and Angelos Kouprizas\authorrefmark{1}}
\affil{Department of Electrical and Computer Engineering, University of Patras, 26504 Patras, Greece}
\corresp{Corresponding author: Eleftherios Mylonas (email: e.mylonas@ac.upatras.gr).}
\authornote{\textbf{This work has been submitted to the IEEE for possible publication. Copyright may be transferred without notice, after which this version may no longer be accessible.} \\This work was funded by the European Union under the Horizon Europe projects: COP-PILOT (Grant Agreement No. 101189819), 3C4AI (Grant Agreement No. 101298630). Views and opinions expressed are however those of the author(s) only and do not necessarily reflect those of the European Union or HADEA. Neither the European Union nor the granting authority can be held responsible for them.}

\begin{abstract}
This paper presents PTQ-NAS-Bench, a multi-scheme Post-Training Quantization (PTQ) extension of the NAS-Bench-201 Neural Architecture Search (NAS) benchmark, and PTQ-CoNAQ, a {{{reference BitOps-constrained architecture-scheme search pipeline with downstream hardware mapping}}}. PTQ-NAS-Bench releases ground-truth quantized {{{accuracy, both after PTQ and two epochs of finetuning, for}}} {up to} 15,625 CIFAR-10 architectures under four PTQ schemes {{with}} 8-bit and sub-8-bit {{weights}} and {{activations}}, obtained through {{a}} 324-configuration {{PTQ}} sweep. PTQ-CoNAQ scores candidates with a zero-cost PTQ sensitivity metric, {{ranks them per scheme, and}} exports the top ten candidates across schemes based on a hardware cost {{{budget for mapping to edge accelerators.}}} Our reference metric, the Quantization Noise Proxy (QNP), attains Spearman correlations of {{$-0.69$, $-0.71$ and $-0.43$}} {over all measured architectures} against measured accuracy at 8-bit, 4-bit and {{{2-bit-weight}}} schemes, while our scheme-selection policy {attains a} mean accuracy regret {of 2.44} points against an oracle at 0.25 GBitOps, against {4.98} for random selection. {{{The overall results reveal that QNP does not improve on FP32 accuracy and BitOps, while a most expensive-feasible scheme policy does better at every budget. These}}} contributions establish a reproducible foundation for hardware-aware NAS under realistic deployment {{{precision, and a baseline that quantization proxies must surpass.}}}
\end{abstract}

\begin{IEEEkeywords}
{Neural Architecture Search, Quantization, Benchmark, Hardware Acceleration, Edge AI}
\end{IEEEkeywords}


\maketitle

\section{INTRODUCTION}
\IEEEPARstart{A}{rtificial} Intelligence (AI) has undergone explosive growth over the last decade, {{and AI services now run}} across the full cloud-edge continuum{{,}} from industrial controllers to conversational assistants. This burst of AI has been especially noticeable in domains like cyber-physical systems, where AI services {{on}} edge controllers {{moderate}} complex procedures across industrial, energy and communication grids. This {{creates a}} need {{for}} cloud-grade {{computation close}} to the edge of the network, near the data sources{{,}} to enable and accommodate these emerging services. {{Most}} of these services run on top of a highly diverse landscape of edge computing engines, each with different software stacks, tool support, hardware features and power profiles, resulting in a persistent lack of uniformity. {{New AI-capable chips increase this diversity further}}, adding considerable complexity for developers who must design, tune and deploy models across hardware {{{platforms \cite{sander-edgeai-survey}.}}} Neural Architecture Search (NAS) \cite{nas-survey} {{reduces this complexity: it}} automates the search and selection of optimized models for a specific task, typically among variants of already-proven architectures{{.}}

However, NAS still faces several open challenges. NAS methods rely heavily on the existence of dedicated benchmarks and supernets for their search --- pools of optimized architectures to choose {{{from, such as NAS-Bench-201 \cite{nas-bench-201}, NATS-Bench \cite{nats-bench}, TransNAS-Bench-101 \cite{transnas-bench-101}, NAS-Bench-360 \cite{nas-bench-360} and the OFA supernet \cite{ofa}}}} . However, the large majority of them reports only full-precision 32-bit (FP32) models and FP32 accuracy results, leaving other computation precisions largely unexplored, {{a}} major hit {{for}} constrained edge {{devices}}. There have been works that try to remedy this by incorporating quantization early in the NAS process, but these methods are {{costly}} since they {{recreate}} existing benchmarks from the {{{beginning, as in JASQ \cite{jasq} and the mixed-precision methods surveyed in \cite{mpnn-survey}}}}. Moreover, no systematic analysis exists of established NAS benchmarks with respect to different quantization schemes and computation precisions, despite quantization being standard practice in any real hardware deployment. Finally, few works integrate architecture search, quantization and hardware mapping into a single joint process directly applicable for hardware accelerators targeting Application Specific Integrated Circuit (ASIC) and Field Programmable Gate Array (FPGA) deployments.

{{{This paper extends our previous work in \cite{mylonas-ieeesocc2026}, which studied the effects of a single 4-bit PTQ scheme. The}}} main contributions of this work are: 
\begin{itemize}
\item {{\textbf{PTQ-NAS-Bench}}}, the first complete extension {{of}} NAS-Bench-201 \cite{nas-bench-201} {{with}} a {{new}} quantization dimension{{, from a}} study {{of}} Post-Training Quantization (PTQ) {{configurations}} to ground-truth {{results under}} four diverse quantization schemes on CIFAR-10 \cite{cifar10}, obtained via the open-source Brevitas \cite{brevitas} toolflow;
\item {{\textbf{PTQ-CoNAQ},}} a new multi-objective NAS procedure for {{{BitOps-constrained architecture-scheme search with downstream hardware}}} mapping{{. It scores architectures with PTQ-aware}} zero-cost proxies {{and supports}} two hardware acceleration design tools, FINN \cite{finn1, finn2} and CGRA4ML \cite{cgra4ml}, {{for}} FPGA/ASIC deployments{{;}}
\item a full characterization of the effects of different PTQ schemes on NAS-Bench-201 architectures, including a multi-scheme Pareto stability analysis across all four quantization schemes, plus a comparison {{of predictor-guided search strategies against ground truth.}}
\end{itemize}

The rest of the paper is structured as follows: Section \ref{background} presents the background {{and}} Section \ref{sota} {{the related work and motivation}}; PTQ-NAS-Bench and PTQ-CoNAQ are presented in Sections \ref{ptq-nas-bench} and \ref{conaq} respectively; Section \ref{experiments} {{reports the}} results, {{{Section \ref{sec:discussion} discusses them and}}} Section \ref{conclusion} concludes the article.

\section{BACKGROUND}\label{background}

\subsection{{NAS \& Hardware-Aware NAS}}
NAS is a subfield of Automated Machine Learning (AutoML) \cite{automl-survey} that {automates} the search {for} optimized neural network architectures. A NAS procedure {combines} a \textit{search space} of candidate architectures (e.g., cell-based spaces with a fixed set of operations and connectivity patterns), a \textit{search strategy} that samples from {it} (evolutionary algorithms, reinforcement learning, etc.), and a \textit{performance estimation strategy} that evaluates or predicts each candidate's quality, most commonly its accuracy. {An \textit{architecture encoding} turns each} candidate into a vector, {which leads} to more accurate prediction. Hardware-aware NAS \cite{b8} {extends} the {search to several} performance metrics $f_i$, {$i\in[1,\ldots,n]$. On} top of a NAS benchmark or a supernet, where {the weights are} fixed per candidate architecture $a$, it {becomes}

\begin{equation}
a^{*}_{NAS}=\arg\max_{a}f_{1}(a),...,f_{n}(a)\label{eq3}
\end{equation}

{usually under} a user-assigned computation budget. {Such} problems are solved by searching for {the} set of solutions {that} no {other solution dominates,} the \textit{Pareto front} {\cite{b8}, where} $a$ {dominates} $b$ $(a>b)$ {if}

\begin{equation}
\forall i \in [1,...,n], f_{i}(a) \geq f_{i}(b) \wedge \exists j f_{j}(a) > f_{j}(b)\label{eq4}
\end{equation}

Pareto rank predictors, {also called} Pareto rank surrogate models, are {performance estimators} trained directly {on} the true ranking of the candidates. {With an} appropriate architecture encoding, {they can separate Pareto-front architectures that differ only slightly} \cite{hw-pr-nas}.

\subsection{Benchmarks \& Supernets in NAS}
\textit{NAS benchmarks} are collections of {precomputed results,} mainly {accuracies, that let a NAS method scan a search space in a fraction} of {the time and} enable reproducible comparison among methods; NAS-Bench-201 \cite{nas-bench-201} and NAS-Bench-360 \cite{nas-bench-360} are well-known {examples. Assembling} a {benchmark, however, requires substantial} effort and computing power. {\textit{NAS supernets} are the} alternative: {because a} search space {is} a known {set} of combinations {of} predetermined {operations} and connectivity patterns, a supernet {contains} all {of them as} subnets and is trained only once. Supernets are usually distributed with APIs that {return} subnets and their associated metrics; OFA \cite{ofa} established {this approach.}

\section{MOTIVATION \& RELATED WORK}\label{sota}

In this section, we present our motivation for organizing PTQ-NAS-Bench and developing PTQ-CoNAQ, along with related work from the literature.

\subsection{Quantization Dimension in Existing NAS Benchmarks}

Hardware-aware NAS has been established nowadays as the de-facto method for AI model search. However, most of the existing hardware-aware NAS benchmarks are limited to only collecting hardware metrics from different platforms. A classic example is the case of HW-NAS-Bench \cite{hw-nas-bench}, which extends NAS-Bench-201 with latency and energy measurements from six hardware platforms, enabling hardware-aware architecture comparison without physical hardware access. All these datasets, although useful, they mostly constraint themselves to FP32 accuracy data or they do not report analytical computation precision information. All the aforementioned result in the lack of  quantization-aware NAS benchmarks, the further increment of FP32 ones and the disablement of any real study of Pareto front stability and perturbation under quantization. Of course, there are some exceptions as is the case with \cite{emq}, which focuses on mixed-precision quantization search and analyzes a specific subset of well-known models. Although useful, it does not apply its methods on existing NAS benchmarks and no real Pareto front study takes place. 

\subsection{Joint Quantization-NAS Methods}

There have been many works in the past that have considered joint {{{quantization-NAS, starting from JASQ \cite{jasq}}}}, a robust methodology that aims to bridge the gap between quantization and NAS. APQ \cite{apq} couples architecture, pruning, and quantization within a single once-for-all supernet; SimQ-NAS \cite{simq-nas} searches sub-network topology and quantization policy simultaneously across CNN and transformer spaces; more recently, QuantNAS \cite{quantnas} searches directly from a fully quantized supernet, with a super-resolution variant using differentiable quantization noise as a surrogate for actual quantization during search. Collectively, these methods achieve great results by expanding the search space, but they suffer from the same problems NAS benchmark papers face, that is, the preparation time and the needed computing power for Quantization-Aware Training (QAT). None of those works attempt to approach joint quantization-NAS in a way that enables the use of already existing NAS benchmarks and supernets. In addition, they assume per-layer mixed precision --- which many tools do not currently support for migration of AI models in real hardware, evaluate on sampled sub-spaces, and never establish ground-truth quantized accuracy over a complete architecture space.

\subsection{From NAS to Custom Silicon}

Hardware-aware NAS works has also sparked a new field of study that centers around the search of not only the best ranking models, but also their most appropriate configurations for hardware acceleration. NASH \cite{nash} selects among candidate operations in hardware-oriented CNNs and automates hardware model generation, reporting accuracy–resource-utilisation Pareto curves; NAAS \cite{naas} proposes a two-level co-search over neural architectures and accelerator configurations, while works like NAHAS \cite{nahas} expand on the accelerator optimization options; AnalogNAS \cite{analognas} targets analog in-memory computing accelerators via hardware-aware training with simulated device noise. Across these works, the numerical precision at which the model will execute is fixed a priori, folded into the search space, or left implicit, and no point on it establishes how the architectural rankings driving the search survive the transition to quantized deployment. 

\subsection{Beyond State-of-the-art}
This work fills these gaps by exploring quantization and specifically PTQ effects across a complete NAS benchmark. It expands the well-known NAS-Bench-201 across four diverse 8-bit and sub-8-bit PTQ schemes after a rigorous study of PTQ methods, obtaining valuable ground-truth data on accuracy, architecture robustness on quantization and recovery analysis. In addition, it leverages the newly obtained data for the creation and testing of a new joint quantization-NAS-hardware mapping pipeline based on PTQ, existing NAS benchmarks and a PTQ noise-aware Pareto rank predictor, managing to export optimized AI hardware accelerators for deployment across the edge.


\section{PTQ-NAS-BENCH: A MULTI-SCHEME QUANTIZATION BENCHMARK FOR HARDWARE-AWARE NAS}\label{ptq-nas-bench}

In this section, we introduce \textbf{PTQ-NAS-Bench}, describing the principle behind its construction, its structure and the coverage and provenance of the data it releases.

\subsection{Core Idea}

PTQ-NAS-Bench addresses the lack of quantization-aware benchmarks by extending NAS-Bench-201 along an orthogonal quantization axis. The architecture set, training protocol, and data splits of NAS-Bench-201 remain unchanged, so that every quantized measurement we release is directly comparable to the FP32 records established in the literature. What we add is a second coordinate, that is, the measured accuracy under a set of Post-Training Quantization (PTQ) schemes. To our knowledge this is the first NAS benchmark to supply PTQ ground truth across a complete search space.

The quantization axis is not sampled arbitrarily. A search space expanded by modern PTQ methods is large and the majority of its points are either dominated in accuracy or not realizable on the hardware targets we consider. We therefore proceed in two stages: an exhaustive characterization of the configuration space on a single representative architecture, followed by a selection of the most promising PTQ schemes, which are then applied to the search space in full. Throughout the first stage, we apply only uniform PTQ: a single weight bitwidth and an activation bitwidth across the whole network, with no QAT and no per-channel granularity. This is a deployability constraint rather than a simplification, and is imposed by the hardware backends we target, where per-layer mixed precision and per-channel scaling is not supported. Every scheme in PTQ-NAS-Bench thus corresponds to a configuration that can actually be synthesized.

\subsection{Benchmark Analysis}

PTQ-NAS-Bench comprises three components: a configuration sweep that maps the PTQ design space on one architecture, a selection of four schemes drawn from that sweep, and the full search-space extension obtained by applying those four schemes to all of NAS-Bench-201. All those components were derived with the use of the Brevitas tool \cite{brevitas}. {{{{The NAS-Bench-201 checkpoints used in this work were obtained from the NATS-Bench topology-search-space CIFAR-10 ones with $seed=777$ at epoch 200 \cite{nats-bench}. PTQ calibration uses the first 10,000 images of the CIFAR-10 training set, and all accuracies are reported on the full 10,000-image CIFAR-10 test set.}}}}

\subsubsection{PTQ Configuration Sweep \& Winners Selection}

We start with the exhaustive analysis of the PTQ scheme space performed on a single fixed {{{architecture, as in \cite{mylonas-ieeesocc2026}.}}} We choose \textit{model\_id 15624} due to its complex cell topology. Holding the architecture fixed isolates the effect of the quantization scheme itself at a cost of a few hundred evaluations, rather than the cost of the full PTQ scheme search space. The sweep spans six axes, listed in the upper block of Table \ref{tab:config-space}; the remaining settings are shown in the lower block. Weight quantization is held symmetric for all our tests due to zero-centred weight distributions. The remaining axes are enumerated exhaustively, giving a total of 324 configurations. Each configuration is evaluated on the CIFAR-10 test set without any finetuning. {{{{Because the four released schemes are chosen on this test accuracy, the choice of scheme is not independent of the test set, although this does not affect the per-architecture measurements produced later. We state this as a limitation in Section~\ref{sec:discussion}.}}}}

\begin{table}
\caption{PTQ configuration axes analysis.}
\label{table}
\setlength{\tabcolsep}{3pt}
\begin{tabular}{|p{100pt}|p{80pt}|}
\hline
\textbf{Swept Axis} & \textbf{Values} \\
\hline
Activation bitwidth & 2, 4, 8 \\
Weight bitwidth & 2, 4, 8 \\
Bias bitwidth & None, 16, 32 \\
Backend & fx, layerwise \\
Equalization & fx, layerwise, None \\
Activation quantizer & symmetric, asymmetric \\
\hline
\textbf{Fixed settings} & \textbf{Values} \\
\hline
Weight quantizer type & symmetric \\
Scale factor type & power-of-two \\
Weight / act.\ granularity & per-tensor \\
Weight calibration & MSE \\
Activation calibration & static, statistics-based \\
Bias correction & {{None}} \\
Graph equalization iterations & 20 \\
BatchNorm handling & merged \\
\hline
\end{tabular}
\label{tab:config-space}
\end{table}

The results of the sweep are summarized in Table \ref{tab:bitwidth-grid}, where you can see each weight/activation bitwidth pair, the best accuracy and the number of usable configurations among the 36 sharing that pair. In general, 221 of the 324 PTQ configurations tested, all collapsed to chance-level accuracy; the median across the sweep is $10.02\%$, against a full-precision reference of $84.19\%$. By careful examination, we can see that at $2$-bit activations the sweep fails uniformly: the best of 108 configurations reaches $11.82\%$, and this holds even at $8$-bit weights. At $8$-bit activations, by contrast, reducing weights from $8$ to $2$ bits costs only $2.84$ points ($84.76\% \rightarrow 81.92\%$). It is therefore clear that activation precision governs the PTQ region. This asymmetry motivates the inclusion of a weight-aggressive, activation-conservative scheme in the final selection. The remaining PTQ axes have minor influence in the final results. \textit{Layerwise} equalization yields the largest number of usable configurations (47 of 108, against 26 for \textit{fx} equalization and 30 for none), and omitting bias quantization outperforms both 16- and 32-bit bias (45 of 108 usable, against 36 and 22 respectively). Backend and activation quantizer type are close to neutral in aggregate. All four schemes selected below consequently omit bias quantization.

\begin{table}
\caption{Best top-1 accuracy (\%) per bitwidth pair, with the number of configurations exceeding $11\%$ out of the 36 sharing that pair.}
\label{table}
\setlength{\tabcolsep}{3pt}
\begin{tabular}{|p{60pt}|p{50pt}|p{50pt}|p{50pt}|}
\hline
& \multicolumn{3}{p{150pt}|}{\textbf{Activation bitwidth}} \\
\hline
\textbf{Weight bitwidth} & 2 & 4 & 8 \\
\hline
2 & 10.00 \; (0/36) & 14.81 \; (3/36) & 81.92 \; (6/36) \\
4 & 10.76 \; (0/36) & 26.22 \; (18/36) & 81.92 \; (26/36) \\
8 & 11.82 \; (2/36) & 41.29 \; (22/36) & \textbf{84.76} \; (26/36) \\
\hline
\end{tabular}
\label{tab:bitwidth-grid}
\end{table}

From the sweep we keep the four schemes achieving the highest test accuracy within their respective bitwidth regimes. These define the schemes carried forward to the full benchmark, listed in Table \ref{tab:schemes} and denoted Q1--Q4.

\begin{table}
\caption{Quantization schemes released in PTQ-NAS-Bench. $b_w$ and $b_a$ denote weight and activation bitwidth{{{{, Acc. the sweep accuracy of model\_id 15624 without finetuning.}}}}}
\label{table}
\setlength{\tabcolsep}{3pt}
\begin{tabular}{|p{50pt}|p{8pt}|p{8pt}|p{30pt}|p{30pt}|p{37pt}|p{31pt}|}
\hline
\textbf{Scheme} & $b_w$ & $b_a$ & \textbf{Backend} & \textbf{Equal.} & \textbf{Act.Q.Type} & \textbf{Acc. (\%)} \\
\hline
\textbf{Q1} (INT8w8a) & 8 & 8 & layerwise & fx & asym & 84.76 \\
\textbf{Q2} (INT4w4a) & 4 & 4 & layerwise & layerwise & asym & 26.22 \\
\textbf{Q3} (INT2w8a) & 2 & 8 & fx & layerwise & sym & 81.92 \\
\textbf{Q4} (INT2w2a) & 2 & 2 & layerwise & layerwise & asym & 10.00 \\
\hline
\end{tabular}
\label{tab:schemes}
\end{table}

\subsubsection{Full Search-Space Extension}

The third and principal component applies the four schemes of Table \ref{tab:schemes} to the 15,625 architectures of the NAS-Bench-201 CIFAR-10 search space. For each architecture--scheme pair we record test accuracy immediately after {{{PTQ}}} and after a uniform two-epoch { quantization-aware} finetuning pass{{{{, where all trainable parameters are updated with Adam (learning rate $10^{-4}$, batch size 512, cross-entropy loss) over the whole CIFAR-10 training set, without data augmentation. We refer to the recovered results as PTQ+FT, which are used in every result except for Table~\ref{tab:finetune} and Figure~\ref{fig:pre_post_finetune}. Coverage differs per measurement: Q1 and Q4 accuracies are complete for all 15,625 architectures, while Q2 and Q3 counts given in Table~\ref{tab:finetune} in Section \ref{experiments}, present gaps. The first reason is that PTQ accuracy was logged under Q2 only from part-way through the Q2 run, giving 3,184 records, while PTQ+FT accuracy is complete. The second reason is that 627 architectures failed to produce a Q3 model due to PTQ errors. Completing the Q2 PTQ records is planned for a second version of the benchmark.}}}} Alongside the accuracies, we release the BitOps required per architecture, computed analytically as $\text{BitOps} = \text{FLOPs} \times b_w \times b_a$, which we use as the sole cost axis. {{{{The released values assume uniform precision. Under the \textit{layerwise} backend, however, Brevitas keeps the stem convolution at 8 bits. Accounting for this raises the BitOps by a median of $2.8\%$ under Q2 and $14.1\%$ (maximum $85.3\%$) under Q4, moving 90 architectures above the 0.25 GBitOps budget. The released workbook and all results use the nominal values.}}}} Additionally, we release for each architecture--scheme pair a quantization-noise proxy score, a surrogate-predicted score, and the rankings each one induces. These are derived quantities and not measurements, and belong to the method rather than to the benchmark; their definition and validation are presented in Section \ref{conaq}. PTQ-NAS-Bench is released through Zenodo \cite{ptq-nas-bench} under an Apache 2.0 license, and its source code is published in Github, at \url{https://github.com/angeloskouprizas/PTQ-NAS-Bench.git}.

\section{PTQ-CONAQ: CO-OPTIMIZATION OF NEURAL ARCHITECTURE AND QUANTIZATION UNDER PTQ}\label{conaq}

In this section we introduce \textbf{PTQ-CoNAQ}, describing the idea behind it, its structure and its operation.

\subsection{PTQ-CoNAQ Overview}

\begin{figure}[t]
\centerline{\includegraphics[width=3.5in]{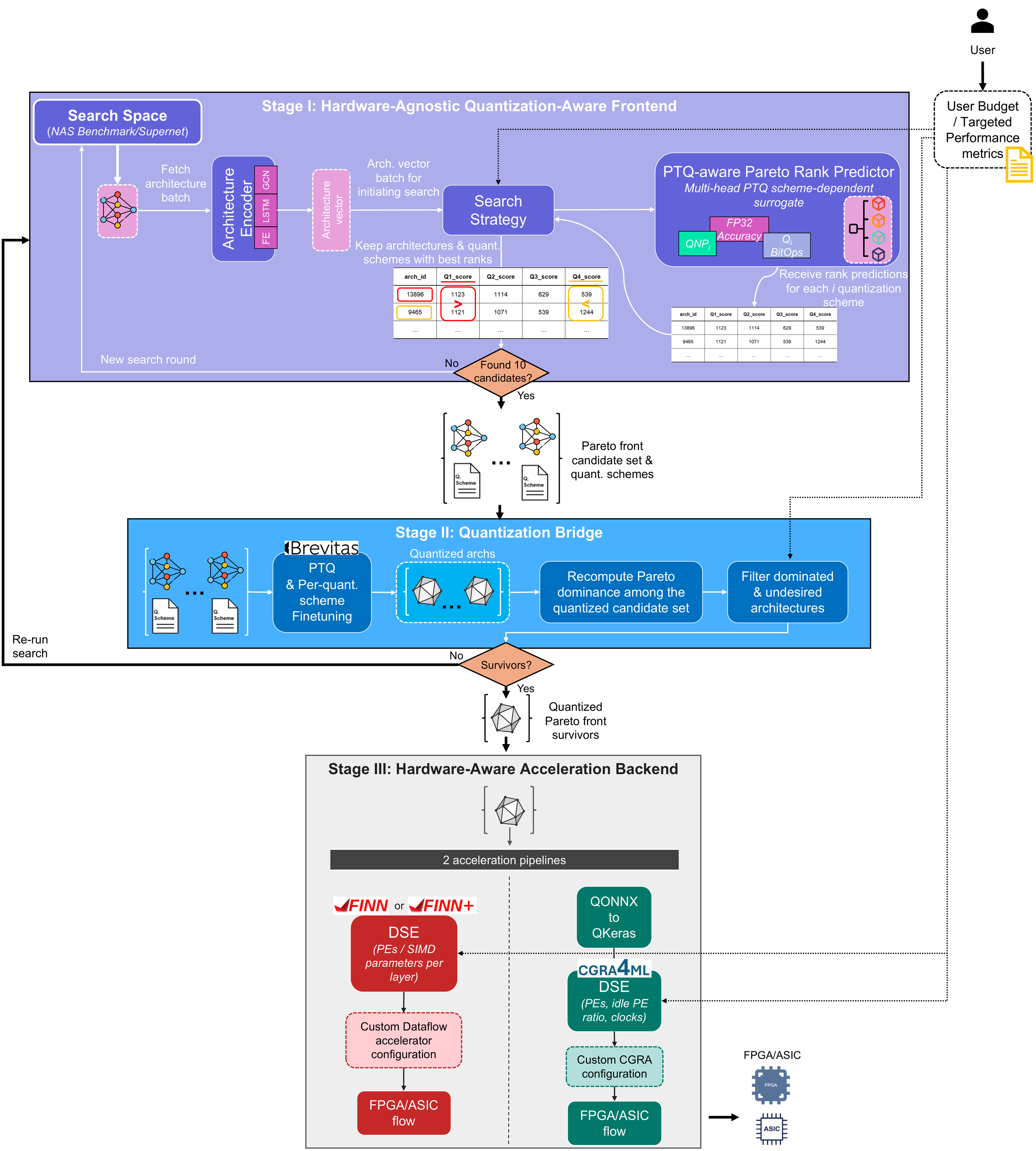}}
\caption{A full view of the proposed PTQ-CoNAQ pipeline.\label{fig:architecture}}
\end{figure}

PTQ-CoNAQ is a new approach to joint quantization-NAS, designed for working with PTQ instead of QAT, FP32 NAS benchmarks and supernets, with support for different FPGA/ASIC flows, for minimizing implementation costs, enabling extensibility and hardware acceleration. The complete PTQ-CoNAQ architecture is depicted in Figure \ref{fig:architecture}. It consists of three independent stages, a hardware-agnostic quantization-aware frontend, a quantization bridge and a hardware-aware backend that are working in sequence, whose role will be explained in the upcoming subsections. PTQ-CoNAQ's source code is also available in PTQ-NAS-Bench Github repository.


\subsection{PTQ Zero-cost Proxies}\label{proxies}

The first major step for PTQ-CoNAQ is to enable the search over the joint architecture--PTQ scheme space with a cheap scoring function. Zero-cost proxies {{{\cite{abdelfattah} and other performance predictors \cite{white2021}} traditionally supply this at {FP32,}}} however, they carry no information about any quantization process. The solution is to have a proxy reflecting the \textit{sensitivity of an architecture to precision reduction}. Such kind of metric has been developed only recently outside main NAS {{{literature, in EMQ \cite{emq} and InfoQ \cite{infoq}}}}. In this work, we restrict PTQ-CoNAQ to a reference proxy instantiation, named \textit{Quantization Noise Proxy (QNP)}. 

QNP is derived from the second-order perturbation metric presented in \cite{hawqv2}, where it is used for selecting candidate mixed-precision bitwidth configurations, and is described as:

\begin{equation}
\Omega = \sum_{\ell=1}^{L} \overline{\mathrm{Tr}}(H_\ell)\,
\big\| Q(W_\ell) - W_\ell \big\|_2^2 ,
\label{eq:omega}
\end{equation}

where $\overline{\mathrm{Tr}}(H_\ell) = \tfrac{1}{n_\ell}\mathrm{Tr}(H_\ell)$ is the average Hessian trace of layer $\ell$, $n_\ell$ is its parameter count, and $\| Q(W_\ell) - W_\ell \|_2^2$ is the squared $\ell_2$ perturbation introduced by quantization. QNP is obtained from \eqref{eq:omega} with {{{{two substitutions and one normalisation, which we describe below:}}}}

\vspace{2pt}
\noindent\textbf{Substitution 1 --- perturbation term.}
For symmetric signed $b$-bit quantization with clipping range $\max|W_\ell|$, the quantization step is

\begin{equation}
\Delta_\ell = \frac{\max|W_\ell|}{2^{\,b-1} - 1}
\label{eq:step}
\end{equation}

Under the high-resolution quantization-noise model \cite{widrow1996}, the rounding error $e = Q(w) - w$ is uniformly distributed on $[-\Delta_\ell/2,\, \Delta_\ell/2]$ and its mean-squared value is the variance of that distribution, $\mathbb{E}[e^2] = \Delta_\ell^2 / 12$. {Summed over the $n_\ell$ elements of the layer, $\| Q(W_\ell) - W_\ell \|_2^2 \approx n_\ell\Delta_\ell^2/12$.} For asymmetric quantization, the same variance is assumed for simplicity. {{{{This model holds when the step is small relative to the range of the quantized signal, although in cases of low bitwidths, the rounding error is no longer negligible.}}}}

\noindent\textbf{Substitution 2 --- curvature term.}
Computing $\mathrm{Tr}(H_\ell)$ requires Hutchinson trace estimation, which, although efficient, still demands repeated Hessian--vector products. {{{We instead use {a gradient-based curvature estimate computed on a single fixed minibatch $\mathcal{B}$ of $N=64$ CIFAR-10 training images. Let $\bar{\mathcal{L}} = \tfrac{1}{N}\sum_{n\in\mathcal{B}}\mathcal{L}_n$ be the mean minibatch cross-entropy loss. The weight curvature term of layer $\ell$ is the squared gradient of $\bar{\mathcal{L}}$, averaged over the $n_\ell$ weights of the layer:}}}}

\begin{equation}F_\ell^{w} = \frac{1}{n_\ell} \sum_{j \in W_\ell}
\left( \frac{\partial \bar{\mathcal{L}}}{\partial w_j} \right)^{2}
\label{eq:fisher-w}
\end{equation}

{{{{This quantity is related to the Hessian through two approximations. First, for a minibatch gradient $\bar{g}_j$ formed from per-example gradients $g_{n,j}$, $\mathbb{E}[\bar{g}_j^{2}] = (\mathbb{E}[g_{n,j}])^{2} + \mathrm{Var}[g_{n,j}]/N$. When the mean training gradient is negligible, as expected for a network trained to convergence as in the case of NAS-Bench-201 models used in this work, this reduces to $\mathbb{E}[g_{n,j}^{2}]/N$, the diagonal of} the empirical Fisher {scaled by $1/N$. Second, the empirical} Fisher approximates the Hessian {only near a minimum and when the model's predictive distribution matches the data \cite{kunstner2019}. Under these conditions, $F_\ell^{w} \approx \overline{\mathrm{Tr}}(H_\ell)/N$, and a single term of \eqref{eq:omega} becomes $\Omega_\ell \approx N\,n_\ell\,F_\ell^{w}\,\Delta_\ell^2/12$. A single minibatch yields one sample of $\bar{g}_j^{2}$, so the estimate is noisy, and architectures far from convergence, including the disconnected cells of NAS-Bench-201, violate the first condition.}}}}

{{{{QNP retains only $F_\ell^{w}\Delta_\ell^2/12$. Dropping $N$ leaves every ranking unchanged, and $n_\ell$ is also dropped, since QNP sums the \textit{per-parameter} expected perturbation of each layer, so each layer contributes according to its average sensitivity.}}}

{{{The activation term follows the same construction, applied to the output $a_\ell$ of every convolutional and fully connected layer over the minibatch, with $m_\ell$ elements in total, $F_\ell^{a} = \frac{1}{m_\ell} \sum_{k \in a_\ell} ( \partial \bar{\mathcal{L}} / \partial a_k )^{2}$,}}}}
{{{{and its step $\Delta_\ell^{a}$ follows \eqref{eq:step}, with the activation bitwidth $b_a$ and the clipping range $\max|a_\ell|$ observed on the minibatch. This term involves three further simplifications. The noise is placed at layer outputs, whereas the quantizers of Section~\ref{ptq-nas-bench} act on layer inputs; the range is the observed maximum, whereas the quantizers are calibrated at the $99.999$th percentile; and the signed symmetric grid of \eqref{eq:step} is used for all schemes, although Q1, Q2 and Q4 quantize activations asymmetrically.}}}} The final form of QNP is therefore:

\begin{equation}
\mathrm{QNP}(A, Q_k) = \sum_{\ell=1}^{L}
\left[
F_\ell^{w} \frac{\Delta_\ell^{w}(b_w)^2}{12} + F_\ell^{a} \frac{\Delta_\ell^{a}(b_a)^2}{12}
\right]
\label{eq:qnp}
\end{equation}

{{{{All four}}} terms are obtained from a single forward--backward pass {{{and require only FP32 weights and a single minibatch of training data in order to operate.}}}} This makes it easy to be deployed both on tabular NAS benchmarks, and equally on subnets sampled from a supernet. {{{Because $b_w$ and $b_a$ enter \eqref{eq:qnp} only through the scaling factor shown in \eqref{eq:step}, QNP's scheme discerning capability is effectively low and discriminates mainly between architectures, so Q1, Q2 and Q4 induce identical architecture rankings.}}} 

\subsection{PTQ-aware Pareto Rank Predictor}\label{predictor}

Having explained the value of PTQ zero-cost proxies, the next step is to use them inside the NAS process. This is done via integration into the Pareto rank predictor architecture. In PTQ-CoNAQ we utilize a multi-head Pareto rank predictor structure, whose heads equal the PTQ schemes Q1--Q4. Each head is trained in rankings produced from combining scheme-specific proxy values, FP32 accuracy and quantized BitOps, which are approximates to the real ones derived from quantized accuracy and BitOps. During operation, the heads receive the same scheme-independent architecture encoding as input, and export scheme-dependent scores. With minimum preparation time for our NAS pipeline in mind, we derive the encoding and training procedures from \cite{hw-pr-nas}, with the only exceptions being the enhancement of the encoder's LSTM submodule with a Pool layer at the output, and the use of 4,000 architectures via Latin HyperCube Sampling for predictor training. {{{{All four heads are trained on the same Latin-hypercube sample. For Q3, the sampled architectures without a Q3 measurement are replaced by measured ones. The remaining architectures form each head's held-out set (11,625 for Q1, Q2 and Q4; 10,998 for Q3), which is used for head admissibility and held-out fidelity in Section~\ref{experiments}.}}}}

\subsection{Stage I: Hardware-agnostic Quantization-Aware Frontend}


PTQ-CoNAQ's first stage is the hardware-agnostic quantization-aware frontend and is built around the PTQ-aware Pareto rank predictor. {It} is {built} to be extensible to any FP32 NAS benchmark or {supernet, and works} on batches of architecture encodings, with its goal being to identify the ten top-score ones and their associated PTQ schemes.

Algorithm \ref{alg:alg1_primitives} defines the three primitives of PTQ-CoNAQ. \textsc{PredictScores} {returns} a score matrix $\mathbf{S}\in\mathbb{R}^{N\times4}$ {for} a batch of $N$ {architectures, with one column per PTQ scheme.} Because each predictor head is trained independently, its scores are not directly comparable across schemes. \textsc{SelectCandidates} is designed around this restriction: it {keeps} the $K\ge10$ highest-scoring architectures \textit{within} each column. Cross-scheme arbitration {is deferred to} \textsc{SelectScheme}, which {min--max normalises the scores over} the full candidate {pool, restricts each architecture to the schemes within} the BitOps budget $C$, and {uses} heads {below} the {fidelity} threshold {$\tau$} only as a fallback. {Min--max normalisation of independently trained heads is a heuristic method and its evaluation is shown in Section~\ref{experiments}. Algorithm \ref{alg:alg2_RS} presents} our customized flavors of the well-known Random Search (RS) and Multi-Objective Evolutionary Algorithm (MOEA) strategies{, which use the same primitives at a cost of} $M$ and {at most $P+T$ predictor evaluations respectively.}

\begin{algorithm}
\caption{PTQ-CoNAQ Primitives}
\label{alg:alg1_primitives}
\begin{algorithmic}
\STATE \textsc{PredictScores}$(\mathbf{E})$
\STATE \hspace{0.5cm}\textbf{Input:} Batch of $N$ architecture encodings $\mathbf{E}$
\STATE \hspace{0.5cm}$\mathcal{Q}\gets\langle\mathrm{INT8w8a},\mathrm{INT4w4a},\mathrm{INT2w8a},\mathrm{INT2w2a}\rangle$
\STATE \hspace{0.5cm}\textbf{for each} $q\in\mathcal{Q}$
\STATE \hspace{1.0cm}$\mathbf{s}_q\gets\mathrm{Head}_q(\mathbf{E})$
\STATE \hspace{0.5cm}\textbf{end for}
\STATE \hspace{0.5cm}\textbf{return}
$\mathbf{S}=[\mathbf{s}_{\mathrm{INT8w8a}},
\mathbf{s}_{\mathrm{INT4w4a}},
\mathbf{s}_{\mathrm{INT2w8a}},
\mathbf{s}_{\mathrm{INT2w2a}}]\in\mathbb{R}^{N\times4}$
\STATE
\STATE \textsc{SelectCandidates}$(\mathbf{S},\mathbf{E},\mathcal{R})$
\STATE \hspace{0.5cm}\textbf{Input:} Predicted score matrix $\mathbf{S}$, encodings $\mathbf{E}$,
running ranges $\mathcal{R}=\{(m_q,M_q)\}_{q\in\mathcal{Q}}$
\STATE \hspace{0.5cm}\textbf{Parameter:} Per-scheme retention size $K\ge10$
\STATE \hspace{0.5cm}\textbf{for each} $q\in\mathcal{Q}$
\STATE \hspace{1.0cm}$m_q\gets\min(m_q,\min_i S_{i,q})$;\quad
$M_q\gets\max(M_q,\max_i S_{i,q})$
\STATE \hspace{1.0cm}$I_q\gets\mathrm{TopK}(\mathbf{s}_q,K)$
\STATE \hspace{0.5cm}\textbf{end for}
\STATE \hspace{0.5cm}\textbf{return}
$\{(e_i,q,S_{i,q})\mid q\in\mathcal{Q},\,i\in I_q\}$, $\mathcal{R}$
\STATE
\STATE \textsc{SelectScheme}$(\mathcal{C},\mathcal{R})$
\STATE \hspace{0.5cm}\textbf{Input:} Candidate triples $\mathcal{C}$,
final per-scheme ranges $\mathcal{R}=\{(m_q,M_q)\}_{q\in\mathcal{Q}}$
\STATE \hspace{0.5cm}\textbf{Parameters:} BitOps budget $C$,
rank-fidelity threshold $\tau$
\STATE \hspace{0.5cm}$\mathcal{Q}_{\mathrm{rel}}\gets\{q\in\mathcal{Q}\mid
|\rho_q|\ge\tau\}$
\STATE \hspace{0.5cm}$\hat{s}_{e,q}\gets(s_{e,q}-m_q)/(M_q-m_q)$
\quad for all $(e,q,s_{e,q})\in\mathcal{C}$
\STATE \hspace{0.5cm}\textbf{for each} distinct architecture $e$ in $\mathcal{C}$
\STATE \hspace{1.0cm}$\mathcal{F}_e\gets\{q\mid(e,q,\cdot)\in\mathcal{C},\
\mathrm{BitOps}(e,q)\le C\}$
\STATE \hspace{1.0cm}\textbf{if} $\mathcal{F}_e=\emptyset$ \textbf{then continue}
\STATE \hspace{1.0cm}\textbf{if} $\mathcal{F}_e\cap\mathcal{Q}_{\mathrm{rel}}
\neq\emptyset$ \textbf{then}
$\mathcal{F}_e\gets\mathcal{F}_e\cap\mathcal{Q}_{\mathrm{rel}}$
\STATE \hspace{1.0cm}$q_e^\star\gets\arg\max_{q\in\mathcal{F}_e}\hat{s}_{e,q}$;
\quad $v_e\gets\hat{s}_{e,q_e^\star}$
\STATE \hspace{0.5cm}\textbf{end for}
\STATE \hspace{0.5cm}$I\gets\mathrm{TopK}(\mathbf{v},10)$
\STATE \hspace{0.5cm}\textbf{return} $\{(e_i,q_i^\star,v_i)\mid i\in I\}$
\end{algorithmic}
\end{algorithm}



\begin{algorithm}
\caption{Batch-processing Predictor-guided RS {and MOEA Strategies}}
\label{alg:alg2_RS}\label{alg:alg3_moea}
\begin{algorithmic}
\STATE \textsc{RS\_Search}$(\mathbf{E})$
\STATE \hspace{0.5cm}\textbf{Input:} Set of $M$ architecture encodings
$\mathbf{E}$ sampled uniformly
\STATE \hspace{0.5cm}\textbf{Parameter:} Batch size $N$
\STATE \hspace{0.5cm}Partition $\mathbf{E}$ into $B=\lceil M/N\rceil$ batches
\STATE \hspace{0.5cm}$\mathcal{C}\gets\emptyset$;\quad
$\mathcal{R}\gets\{(+\infty,-\infty)\}_{q\in\mathcal{Q}}$
\STATE \hspace{0.5cm}\textbf{for each} batch $\mathbf{E}_b$
\STATE \hspace{1cm}$\mathbf{S}\gets\textsc{PredictScores}(\mathbf{E}_b)$
\STATE \hspace{1cm}$\mathcal{C}_b,\mathcal{R}\gets\textsc{SelectCandidates}(\mathbf{S},\mathbf{E}_b,\mathcal{R})$
\STATE \hspace{1cm}$\mathcal{C}\gets\mathcal{C}\cup\mathcal{C}_b$
\STATE \hspace{0.5cm}\textbf{end for}
\STATE \hspace{0.5cm}$\mathcal{C}_{10}\gets\textsc{SelectScheme}(\mathcal{C},\mathcal{R})$
\STATE \hspace{0.5cm}\textbf{return} $\mathcal{C}_{10}$
\STATE
\STATE \textsc{MOEA\_Search}{$(\mathcal{A})$}
\STATE \hspace{0.5cm}{{{\textbf{Input:} {Search space $\mathcal{A}$}}}}
\STATE {{\hspace{0.5cm}}}{\textbf{Parameters:} Population size $P$, {{{mutation steps $T$, tournament size $k$}}}}
\STATE \hspace{0.5cm}\textbf{for each} $q\in\mathcal{Q}$
\STATE {{\hspace{1cm}{{$\mathbf{P}_q\gets P$ distinct architectures sampled uniformly from $\mathcal{A}$, scored by $\mathrm{Head}_q$}}}}
\STATE {{\hspace{1cm}{\textbf{for} {$t=1,\ldots,T$}}}}
\STATE {{\hspace{1.5cm}{{$p\gets$ best of $k$ individuals drawn from $\mathbf{P}_q$}}}}
\STATE {{\hspace{1.5cm}{{$o\gets\mathrm{Mutate}(p)$}}}}
\STATE {{\hspace{1.5cm}{{\textbf{if} $o$ is valid and $o\notin\mathbf{P}_q$ \textbf{then} $s_o\gets\mathrm{Head}_q(o)$}}}}
\STATE {{{\hspace{1.5cm}{\textbf{else} $s_o\gets-\infty$}}}}
\STATE {{{\hspace{1.5cm}{$\mathbf{P}_q\gets\mathbf{P}_q\cup\{o\}$ minus its lowest-scoring individual}}}}
\STATE {{\hspace{1cm}}}{{{\textbf{end for}}}}
\STATE \hspace{1cm}$\mathcal{C}_q\gets\mathrm{TopK}(\mathbf{P}_q,10)$
\STATE \hspace{0.5cm}\textbf{end for}
\STATE \hspace{0.5cm}$\mathcal{C}\gets\bigcup_{q\in\mathcal{Q}}
\{(e,q,\mathrm{Head}_q(e))\mid e\in\mathcal{C}_q\}$
\STATE \hspace{0.5cm}$\mathcal{C}_{10}\gets\textsc{SelectScheme}(\mathcal{C},\mathcal{R})$
\STATE \hspace{0.5cm}\textbf{return} $\mathcal{C}_{10}$
\end{algorithmic}
\end{algorithm}

\subsection{Stage II: Quantization Bridge}

PTQ-CoNAQ's second stage is the quantization bridge, and can be assumed as a middleware stage. Its purpose is the quantization of the top ten candidate architectures and their final evaluation before hardware implementation. The first step is the application of the winning PTQ schemes on their respective AI models. At this point, a scheme-dependent recovery finetuning can be applied. For Q1 no finetuning is applied, for Q2 and Q3 2 epochs are enough{ \cite{mylonas-ieeesocc2026}}, whereas Q4 requires at least 25 epochs for best architectures to recover. {On a sample of 150 architectures, 25 epochs raise Q4 accuracy from 26.4\% to 50.5\% on average, and it is still rising at epoch 25, while resulting in architecture reordering (Kendall $\tau=0.52$ between the 2- and 25-epoch rankings).}The next step is to rank the final quantized models based on ground-truth data and filter the dominated and undesired ones based on the cost budget. The surviving models are forwarded to the final stage. All PTQ and finetuning tasks run again with the AMD-Xilinx's open-source Brevitas tool. Lastly, a feedback loop to Stage I exists in case no survivors are left, for which a new search is triggered.

\subsection{Stage III: Hardware-aware Acceleration Backend}

PTQ-CoNAQ's third stage is the hardware-aware acceleration backend and is responsible for the optimal mapping of the survivor models to customized hardware accelerators for FPGAs/ASICs. It supports two types of hardware acceleration flow, one for high-performance dataflow accelerators based on FINN, and the other for low-power Coarse-Grained Reconfigurable Array (CGRA)-based designs with CGRA4ML. Currently, the design decision lies with the user who chooses which flow to follow at the beginning of the PTQ-CoNAQ search. PTQ-CoNAQ also allows the user to test a set of quantized models on either flow without the need for restarting, which allows for easy comparison among them. At this stage, the uniform PTQ decision we took in the beginning pays back, since neither tool allows per-layer mixed-precision yet. For enabling optimal hardware mapping, we leverage the utilities provided from each tool for Domain Space Exploration (DSE). The AI model translation routine for CGRA4ML and the corresponding DSE process is inherited {{{{from our previous work \cite{mylonas-ieeesocc2026}}}}}, while for FINN we utilize the existing \textit{step\_target\_fps\_parallelization} routine for greedy search in the hardware search space. 

\section{EXPERIMENTATION}\label{experiments}

In this section, we present the results of our experiments and provide a summary of our findings. {{We first characterize the benchmark itself, and then evaluate PTQ-CoNAQ and its hardware mapping.}}

\subsection{Benchmark Characterization}\label{sec:characterization}

Recording both accuracies is deliberate, since we are able to understand how much information encoded in the FP32 weights survives the quantization and how easily it can be recovered. Table \ref{tab:finetune} shows that the gap between them is strongly scheme-dependent. Q1 is essentially converged, gaining under one point from finetuning. Q2 and Q3 recover by roughly $25$ and $28$ points respectively, with almost every architecture improving. Q4 improves by $10$ points on average, which transforms the distribution --- the fraction of architectures sitting at the chance-level floor falls from $82.0\%$ to {$8.7\%$} --- without bringing the scheme into a competitive accuracy range.

\begin{table}
\caption{Mean test accuracy (\%) before and after two-epoch finetuning,
over {{{{the paired subset of architectures with both measurements}}} for the given {{{scheme}}}}{{{.}}} ``Floor'' is the
fraction of architectures at or below $11\%$ top-1 before {/ after} finetuning.}
\label{table}
\setlength{\tabcolsep}{3pt}
\begin{tabular}{|p{50pt}|p{22pt}|p{22pt}|p{22pt}|p{27pt}|p{60pt}|}
\hline
\textbf{Scheme} & $n$ & \textbf{Pre} & \textbf{Post} & $\Delta$ & \textbf{Floor} { \textbf{(pre / post)}} \\
\hline
\textbf{Q1} (INT8w8a) & 15,625 & 85.40 & 86.18 & $+0.79$ & 0.022 { / 0.022} \\
\textbf{Q2} (INT4w4a) & 3,184 & 56.79 & 81.63 & $+24.84$ & 0.040 { / 0.021} \\
\textbf{Q3} (INT2w8a) & 14,998 & 38.23 & 66.52 & $+28.29$ & 0.366 { / 0.048} \\
\textbf{Q4} (INT2w2a) & 15,625 & 10.38 & 20.40 & $+10.01$ & 0.820 { / 0.087} \\
\hline
\end{tabular}
\label{tab:finetune}
\end{table}

\begin{figure}
\centerline{\includegraphics[width=3.5in]{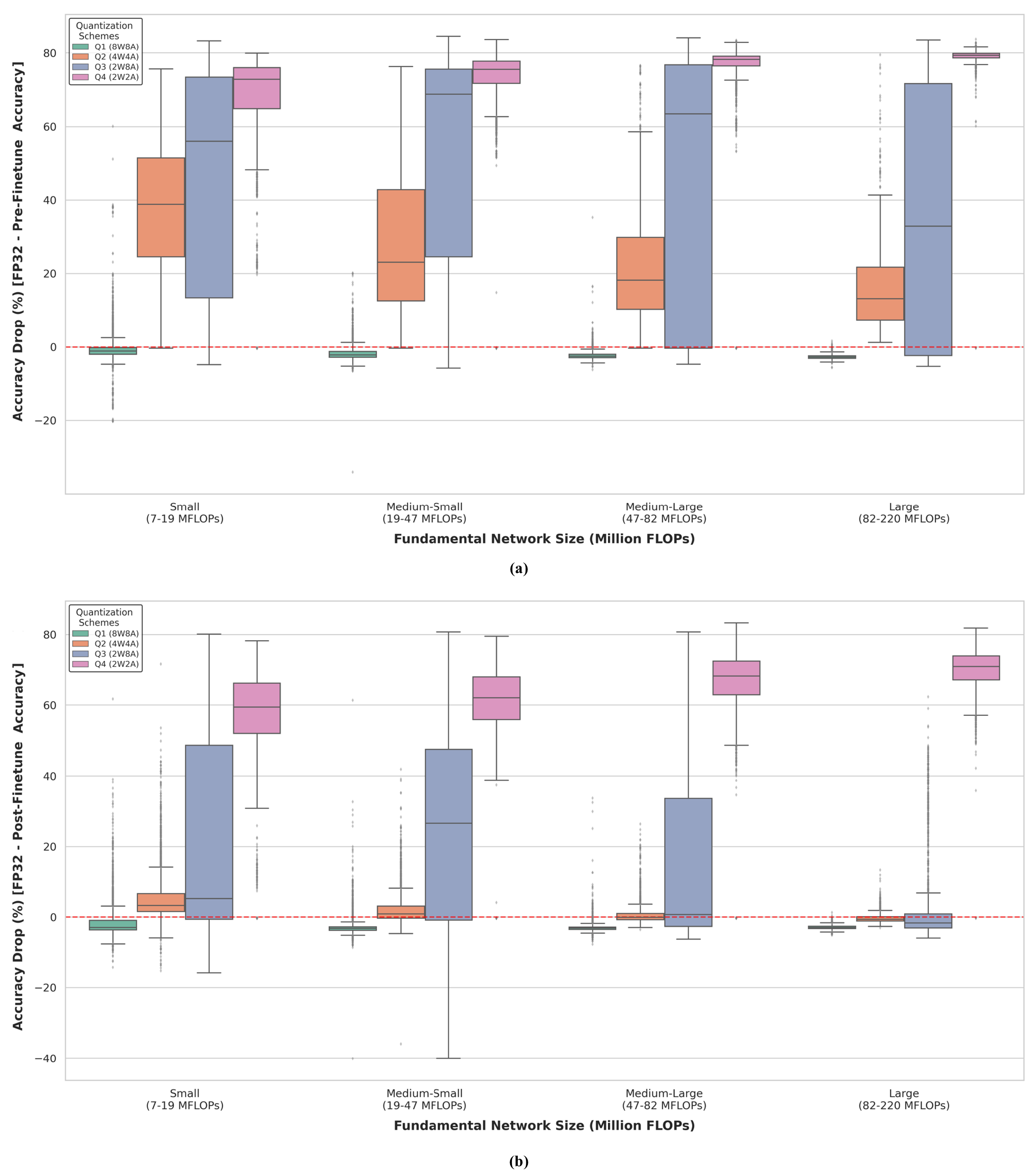}}
\caption{Accuracy Drop boxplots per quantization scheme before and after a two-epoch finetuning. \label{fig:pre_post_finetune}}
\end{figure}

Aggregate means conceal a systematic dependence on network size, which is presented in Figure \ref{fig:pre_post_finetune}. Two opposing trends are visible. Under Q2 and Q3, which reduce weight precision while retaining moderate or high activation precision, larger networks are more robust. The median Q2 drop falls monotonically from $39.3$ points to $13.3$ before finetuning, and after finetuning it recovers almost completely, with a median drop of $-0.5$ points and an interquartile range of $1.2$. Q3 behaves similarly in the largest bin, where the post-finetuning median drop reaches $-1.0$. Under Q4 the trend reverses. The median drop \textit{increases} with scale, from $71.7$ points to $79.1$ before finetuning, and the interquartile range contracts from $12.9$ to $1.6$. Larger networks are not only no better under $2$-bit activations, but they fail more completely and more uniformly. 

These trends are best read as a property of operation composition rather than of {{{scale, as first observed in \cite{mylonas-ieeesocc2026}.}}} Within NAS-Bench-201, FLOPs are almost indistinguishable from the number of \textit{nor\_conv\_3x3} operations with a Pearson correlation of $r = 0.994$, while every remaining operation type (\textit{none}, \textit{skip\_connect}, \textit{avg\_pool\_3x3} and \textit{nor\_conv\_1x1}) exhibit negative correlation. {{{{One possible interpretation}}} is that a $3\!\times\!3$ kernel, which carries nine times the weights of a $1\!\times\!1$ kernel per channel pair, {{{averages the rounding error of individual weights over more terms, so that convolution-rich architectures would be}}} better conditioned under the weight-aggressive schemes Q2 and Q3.} The parameter-free operations that dominate the low-FLOPs region propagate activations without contributing quantizable weights. Under Q4 they offer no compensating weight capacity while still transmitting a $2$-bit-degraded signal, and the apparent advantage of small architectures in that regime reflects a lower FP32 baseline rather than genuine robustness. This is consistent with the sweep results: once activation precision is insufficient to propagate information, the failure saturates. Finetuning shifts the Q4 distribution by roughly $10$ points at every scale without altering the ordering.

A further observation concerns Q3's distribution. Q3's interquartile range exceeds $45$ points in most cases both before and after finetuning, far wider than any other scheme. The Q3 population is essentially split into a substantial subset that tolerates $2$-bit weights with little loss and another that collapses. This bimodality is precisely the structure an architecture-level predictor must capture.

\subsection{QNP Analysis}

\begin{figure}[b]
\centerline{\includegraphics[width=3.5in]{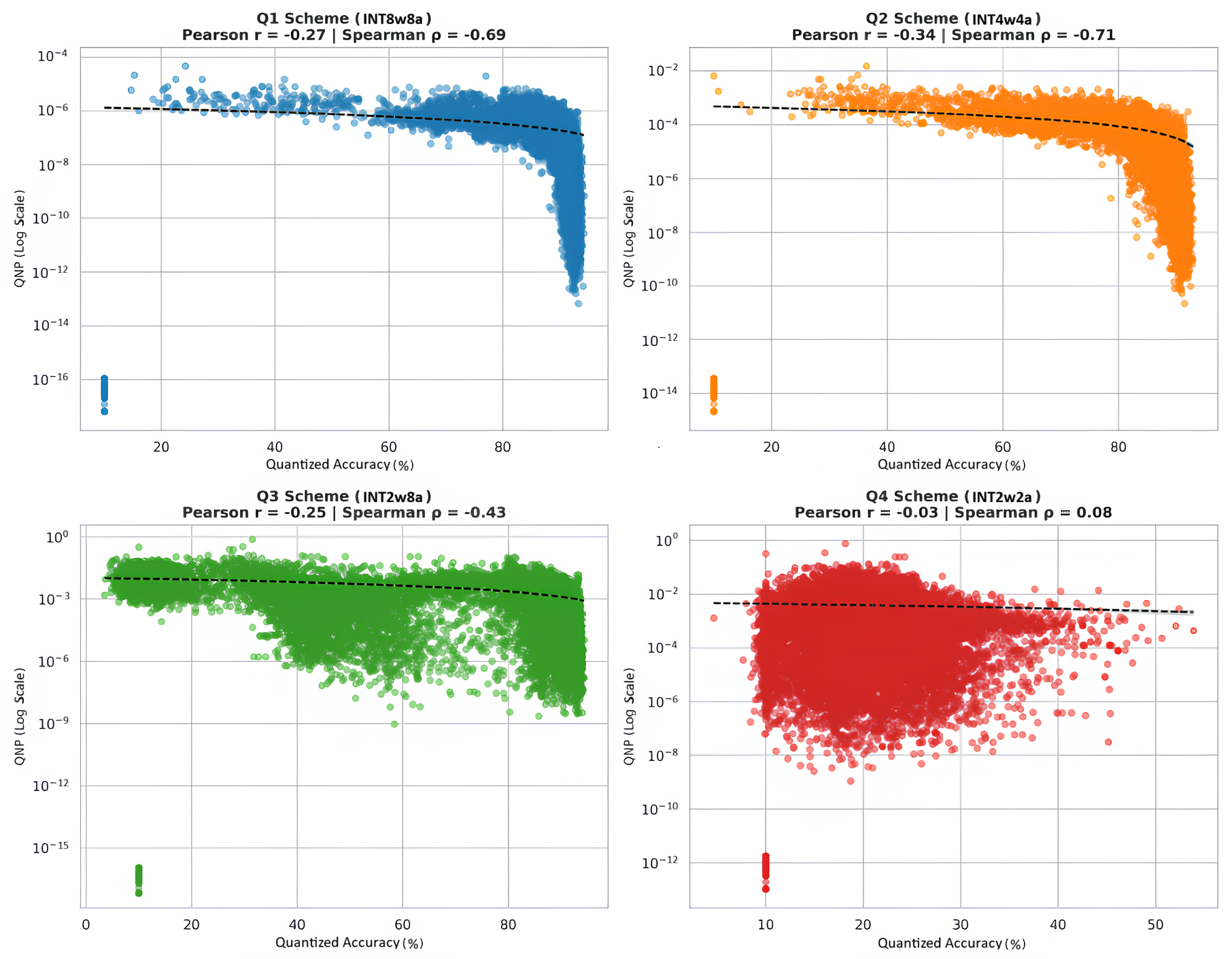}}
\caption{QNP versus quantized accuracy after finetuning under each PTQ scheme. QNP is shown on a logarithmic axis.\label{fig:qnp-scatter}}
\end{figure}

We first start by characterizing our reference quantization noise proxy directly against measured quantized accuracies from PTQ-NAS-Bench. Figure \ref{fig:qnp-scatter} shows negative rank correlation for {{{Q1--Q3 {over all architectures with PTQ+FT accuracy (15,625; 14,998 for Q3)},}}} specifically $\rho=-0.69$ for Q1, $\rho=-0.71$ for Q2 and $\rho=-0.43$ for Q3, which is desirable for a perturbation metric, while $\rho=+0.08$ for Q4. In addition, we see a cluster of architectures at the $9.71\%$ chance-accuracy floor, mistakenly sorted as high-quality ones, which are the 341 architectures whose cell topology provides no path from input to output. Excluding them raises the correlation to $\rho=-0.81$, $-0.83$ and $-0.52$ at Q1--Q3 {{{respectively{, and gives $\rho=+0.03$ at Q4. Unless stated otherwise, all results below use the full population, including the disconnected cells}.}}} The Q4 inversion is not attributable to this cluster but is a failure of the PTQ scheme, which both compresses and reorders the landscape in a way that our two-epoch finetuning is not making much difference. The reversal itself is an empirical finding of the benchmark, independent of any proxy. In addition to this, we report the agreement between proxy-generated and ground-truth Pareto rankings in Table \ref{tab:proxy-fidelity}, before any predictor training. Here we use $\ln(\mathrm{QNP})$ instead of QNP for numerical stability. The results show a stable but declining win of the proxy-free configuration during Q1--Q3, as was expected by \cite{mylonas-ieeesocc2026}, and a total collapse at Q4. This simple outcome justifies the need for PTQ-specific benchmarks. {{{{Adding QNP lowers the fidelity at Q1--Q3, and this is not caused by the disconnected cells: without them, the proxy-free objective still leads with $\tau=0.739$, $0.657$ and $0.378$, against $0.699$, $0.640$ and $0.362$ with QNP. Only} at Q4 {does}}}} its composite $\ln(\mathrm{QNP})\!\times\!\mathrm{FP32\_acc}$ form {{{{lead}, reaching}}} $\tau=0.1112$ against $0.0145$. This outcome reinforces the idea of improving our reference PTQ noise proxy and applying more zero-cost quantization proxies to the test.

\begin{table}
\caption{Rank fidelity (Kendall's $\tau$) between proxy-generated and
ground-truth Pareto rankings, prior to predictor training. All
configurations include quantized BitOps as an objective.}
\label{table}
\setlength{\tabcolsep}{3pt}
\begin{tabular}{|p{85pt}|p{25pt}|p{25pt}|p{25pt}|p{30pt}|}
\hline
\textbf{Objective set} & \textbf{Q1} & \textbf{Q2} & \textbf{Q3} & \textbf{Q4} \\
\hline
$\ln(1{+}\mathrm{QNP})$, FP32\_acc  & 0.6313 & 0.5770 & 0.3171 & $-0.0716$ \\
$\ln(\mathrm{QNP})\!\times\!\mathrm{FP32\_acc}$  & 0.4908 & 0.5212 & 0.2777 & \textbf{0.1112} \\
FP32\_acc only (\textit{no proxy})  & \textbf{0.7418} & \textbf{0.6655} & \textbf{0.3942} & 0.0145 \\
\hline
\end{tabular}
\label{tab:proxy-fidelity}
\end{table}

\subsection{PTQ-CoNAQ Performance Results}

\begin{figure}
\centerline{\includegraphics[width=3.5in]{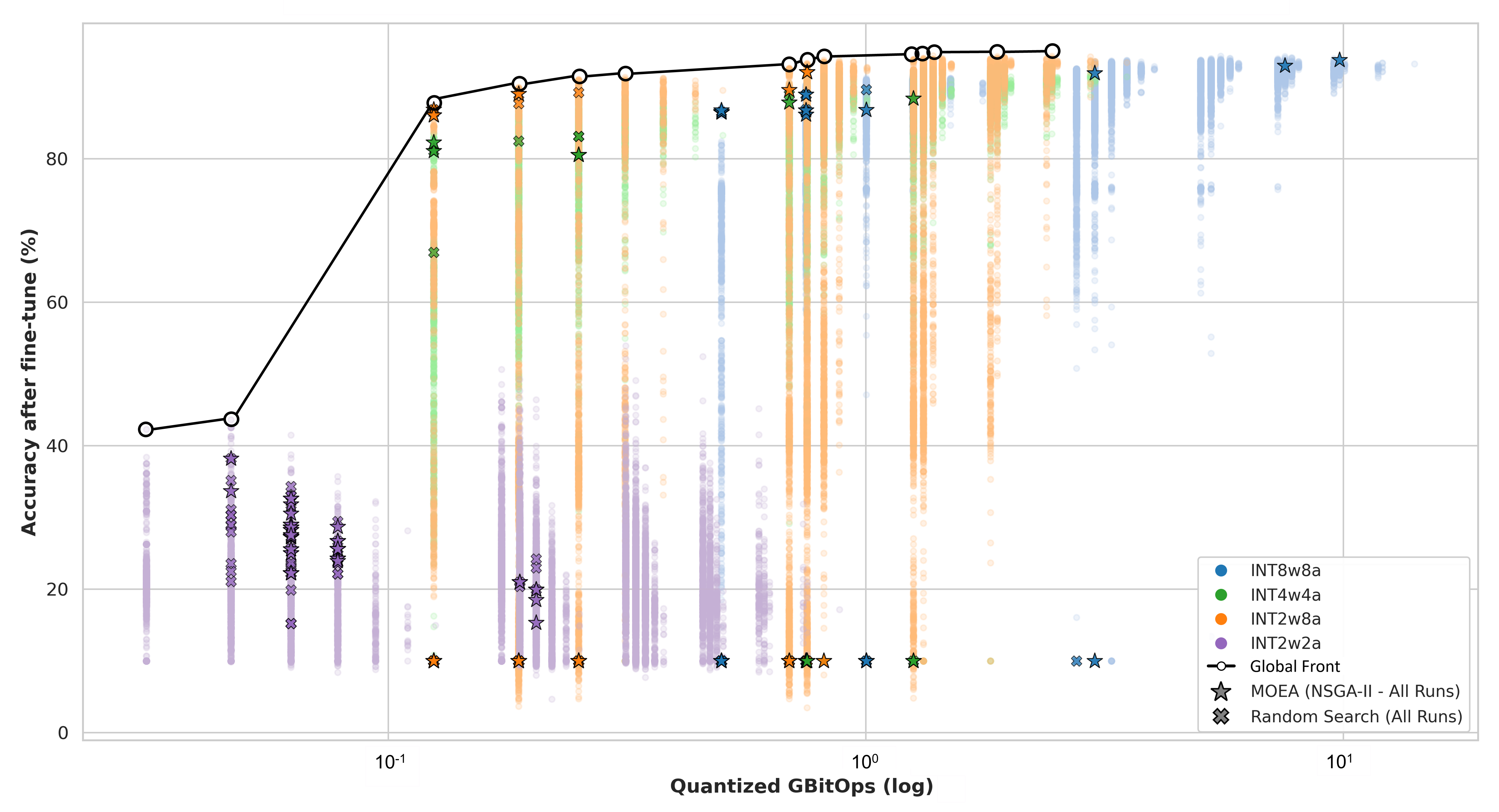}}
\caption{Unified Pareto Space across all PTQ schemes. \label{fig:pareto}}
\end{figure}

We now present PTQ-CoNAQ's results for our QNP-based implementation, using Table \ref{tab:proxy-fidelity}'s first objective set for Q1--Q3 predictor head training and the second one for Q4. {{{Because QNP induces almost the same architecture ranking under Q1--Q3 (Section~\ref{conaq}-\ref{proxies}), these targets carry little scheme-specific information, while discrimination comes mainly from the scheme-dependent BitOps budget and the separate Q4 target.}}} Figure \ref{fig:pareto} presents the unified Pareto space together with the candidates returned by RS and MOEA across $50$ runs per PTQ. The front comprises the best accuracy and budget-friendly options that lie mostly on the center of the figure at Q2--Q3, while it also extends to the remaining high-quality architectures in Q4 region at $\approx44\%$ accuracy. One of the best architectures is found at the edge of Q3 to Q4 region near $0.12$ GBitOps, where Q3 attains roughly $88\%$. Both searches are able to track the Pareto front line at the same level of consistency. Two regions of the returned sets warrant comment. Under Q4 both methods return solutions between $15\%$ and $38\%$, which reflects the compression of the Q4 landscape. In addition, both methods also return a small number of chance-accuracy solutions at every scheme; these are the topologically disconnected architectures as identified above, which are non-dominated in BitOps and in the proxy objective and are mistakenly selected. However, these are trivially removable by a connectivity check on the cell topology prior to search.

Table \ref{tab:fidelity} reports the correlation between each predictor head's ranking and the ground-truth Pareto {{{ranking {(non-dominated fronts of PTQ+FT accuracy and BitOps), computed over the whole dataset, including the 4,000 training architectures}.}}} Kendall's $\tau_b$ is the primary statistic, while Spearman's $\rho$ is reported alongside. We see that Q1--Q3 exhibit comparable fidelity ($\tau_b$ between $0.30$ and $0.33$), while Q4 degrades to $0.158${{{,}}} {{{{consistently}}} with the behaviour established above. {{{On each head's held-out architectures alone the picture is the same: $\tau_b$ is $0.32$ and $0.36$ for Q1 and Q3, $0.24$ for Q2 and $0.15$ for Q4.}}}} While these values are modest in absolute terms, the operative requirement for the downstream policy is that search points to candidates from low non-domination fronts rather than that global rank be recovered exactly. As shown in {Table \ref{tab:rs_vs_moea}}, the ten candidates returned by each search concentrate in the low-front region across all four {{{schemes}}}{{{{. Since a front index is not a percentile, we convert it: the}}} median front of the {{{returned candidates contains, together with all better fronts, the best $17$--$28\%$ of architectures for Q1--Q3 and $27$--$33\%$ for Q4,}}}} with MOEA ahead of RS in median front index throughout.

\begin{table}
\caption{Ranking Consistency between predicted \& ground-truth rankings, after predictor {{training{{, over the whole dataset and over each head's held-out architectures (11,625 for Q1, Q2 and Q4; 10,998 for Q3)}.}}}}
\label{table}
\setlength{\tabcolsep}{3pt}
\begin{tabular}{|p{62pt}|p{34pt}|p{34pt}|p{34pt}|p{34pt}|}
\hline
 & \multicolumn{2}{c|}{{{{\textbf{Whole dataset}}}}} & \multicolumn{2}{c|}{{{{{\textbf{Held-out}}}}}} \\
\hline
\textbf{Predictor head} & {\textbf{{{$\tau_b$}}}} & {\textbf{{{$\rho$}}}} & {{{\textbf{{$\tau_b$}}}}} & {{\textbf{{{$\rho$}}}}} \\
\hline
\textbf{Q1} (INT8w8a) & 0.3184 & 0.4443 & {{{{0.3223}}}} & {{{{0.4660}}}} \\
\hline
\textbf{Q2} (INT4w4a) & 0.3007 & 0.4181 & {{{{0.2411}}}} & {{{{0.3512}}}} \\
\hline
\textbf{Q3} (INT2w8a) & 0.3340 & 0.4750 & {{{{0.3596}}}} & {{{{0.5143}}}} \\
\hline
\textbf{Q4} (INT2w2a) & 0.1581 & 0.2432 & {{{{0.1516}}}} & {{{{0.2249}}}} \\
\hline
\end{tabular}
\label{tab:fidelity}
\end{table}


Table \ref{tab:rs_vs_moea} compares the two search strategies {{{{per scheme, without \textsc{SelectScheme}, over 50 runs. Both are given}}} a budget of 100 {{{search steps: RS scores 100 random architectures and keeps the top ten, while MOEA uses $P=20$, $T=80$ and $k=5$. Rejected offspring (cells with fewer than two convolutions) consume a step without a predictor call, so MOEA performs at most 100 predictor evaluations and in practice fewer.}}}} RS outperforms MOEA on every scheme and on all three metrics, reaching higher accuracy at lower GBitOps and attaining a normalized hypervolume larger by $0.18$--$0.22$ at Q1--Q3. The gap narrows to $0.025$ at Q4, where the compressed landscape leaves little for either strategy to exploit. We attribute this to the {{{search budget of 80 mutations from a population of 20 with possible offspring rejection, exploring a narrower neighbourhood than 100 independent samples.}}} Evolutionary search is therefore not justified at this budget, and the surrogate-guided pipeline delivers its benefit through the predictor rather than through the choice of optimizer.

\begin{table}
\caption{Search performance under RS / MOEA strategies{. Front: median global non-dominated front index of the returned top-10 candidates, with the share of architectures in that front or better.}}
\label{table}
\setlength{\tabcolsep}{2pt}
\begin{tabular}{|p{40pt}|p{30pt}|p{30pt}|p{30pt}|p{36pt}|p{50pt}|}
\hline
\textbf{Scheme} & \textbf{Strategy} & \textbf{Acc (\%)} & \textbf{GBitOps} & \textbf{Norm. HV} & {\textbf{Front (top \%)}}\\
\hline
\textbf{Q1}$^{\mathrm{1}}$ & \textit{RS} & 93.41 & 7.062 & 0.822 & {$\approx$120 (27.9)} \\
 & \textit{MOEA} & 91.66 & 7.641 & 0.640 & {$\approx$100 (20.7)} \\
\hline
\textbf{Q2}$^{\mathrm{2}}$ & \textit{RS} & 91.88 & 1.885 & 0.773 & {$\approx$150 (26.0)} \\
 & \textit{MOEA} & 89.78 & 2.074 & 0.557 & {$\approx$110 (17.0)} \\
\hline
\textbf{Q3}$^{\mathrm{3}}$ & \textit{RS} & 93.42 & 1.971 & 0.796 & {$\approx$180 (24.7)} \\
 & \textit{MOEA} & 92.43 & 2.076 & 0.616 & {$\approx$165 (23.1)} \\
\hline
\textbf{Q4}$^{\mathrm{4}}$ & \textit{RS} & 33.23 & 0.110 & 0.310 & {$\approx$410 (32.5)} \\
 & \textit{MOEA} & 31.14 & 0.123 & 0.285 & {$\approx$345 (26.8)} \\
\hline
\multicolumn{6}{p{222pt}}{$^{\mathrm{1}}$Reference point @ (0.0,14.088). $^{\mathrm{2}}$Reference point @ (0.0,3.522). $^{\mathrm{3}}$Reference point @ (0.0,3.522). $^{\mathrm{4}}$Reference point @ (0.0,0.880).}\\
\end{tabular}
\label{tab:rs_vs_moea}
\end{table}


Table \ref{tab:scheme_policy} compares SelectScheme against fixed-scheme and naive policies at three budgets. {It evaluates the choice of scheme for a given architecture, over every architecture with complete data across the four schemes (14,998; at $0.25$ GBitOps, the 9,722 with at least one feasible scheme). The population includes the 4,000 predictor-training architectures, so the comparison is in-sample. In \textsc{SelectScheme}, $s_{e,q}$ is the predicted rank of $e$ under $\mathrm{Head}_q$, reversed and min--max normalised over the 14,998 architectures. The random policy draws uniformly from the feasible set.} Head admissibility is determined by rank fidelity on {each head's held-out architectures (Table~\ref{tab:fidelity});} a threshold of $0.35$ admits Q1--Q3 to the argmax comparison and flags Q4 as a fallback, which preserves coverage at tight budgets. {Regret is the mean accuracy gap to the oracle, conditional on feasibility; Regret$_0$ scores infeasible choices as zero.} Our policy attains mean regret of {$2.44$, $6.73$ and $8.67$} accuracy points against the oracle at $0.25$, $1.0$ and $4.0$ GBitOps, against {$4.98/17.57/24.60$} for random and {$9.75/39.67/65.62$} for always-Q4.

{However, cost-ordered rules outperform it at every budget. Since Q2 and Q3 have identical BitOps and Q1 is more expensive, Q4 is the only feasible scheme, and also the oracle choice, whenever Q2 is infeasible. ``Q2 if feasible, else Q4'' therefore reaches $0.70/2.12/4.33$, and choosing the most expensive feasible scheme (Q1, then Q2, then Q4) reaches $0.70/1.15/0.72$. Feeding the same rule in other per-scheme rankings locates the weakness: the QNP-based targets on which our heads are trained give $0.72/2.77/4.65$, QNP-free targets (FP32 accuracy and BitOps) reproduce the most-expensive-feasible choices, and even ground-truth per-scheme fronts give $0.98/2.02/2.57$. QNP therefore does not improve scheme selection, and min--max normalised ranks carry no common accuracy scale across schemes. In PTQ-NAS-Bench, a cost-ordered rule is thus a strong} baseline that a {scheme-selection method must surpass.} Finally, the oracle composition shifts from {$79\%$} Q4 at the tightest budget to $55\%$ Q1 at the loosest, confirming that no fixed scheme is adequate across the range.

\begin{table}
\caption{SelectScheme(.) Policy Comparison{{{{, over the 14,998 feasible architectures. Acc and GBitOps are averaged over the architectures where the policy is feasible.}}}}}
\label{table}
\setlength{\tabcolsep}{2pt}
\begin{tabular}{|p{40pt}|p{24pt}|p{26pt}|p{29pt}|p{21pt}|p{25pt}|p{50pt}|}
\hline
\textbf{Policy} & \textbf{Valid\%} & \textbf{Acc(\%)} & \textbf{GBitOps} & \textbf{Regret} & {{{{\textbf{Regret$_0$}}}}} & \textbf{Q1/Q2/Q3/Q4 (\%)} \\
\hline
\multicolumn{7}{|p{230pt}|}{\textbf{C = 0.25 GBitOps} — {{9,722}} feasible architectures} \\
\hline
Oracle & 100.0 & {{30.15}} & 0.164 & 0.00 & {{0.00}} & {{0/11/9/79}} \\
always-Q2 & {{20.6}} & {{65.22}} & 0.167 & {{3.41}} & {{16.75}} & 0/100/0/0 \\
always-Q3 & {{20.6}} & {{48.78}} & 0.167 & {{19.84}} & {{20.13}} & 0/0/100/0 \\
always-Q4 & 100.0 & {{20.40}} & 0.138 & {{9.75}} & {{9.75}} & 0/0/0/100 \\
random & 100.0 & {{25.17}} & 0.155 & {{4.98}} & {{4.98}} & 0/6/7/87 \\
{{most exp.}} & {{100.0}} & {{\textbf{29.45}}} & {{0.164}} & {{\textbf{0.70}}} & {{\textbf{0.70}}} & {{0/21/0/79}} \\
\textit{ours} & 100.0 & {{27.71}} & 0.164 & {{2.44}} & {{2.44}} & {{0/8/12/79}} \\
\hline
\multicolumn{7}{|p{230pt}|}{\textbf{C = 1.0 GBitOps} — {{14,998}} feasible architectures} \\
\hline
Oracle & 100.0 & {{60.03}} & 0.546 & 0.00 & {{0.00}} & {{10/30/24/35}} \\
always-Q1 & 13.3 & {{72.47}} & {{0.668}} & 0.84 & {{50.38}} & 100/0/0/0 \\
always-Q2 & {{64.8}} & {{78.33}} & 0.554 & 3.27 & {{9.26}} & 0/100/0/0 \\
always-Q3 & {{64.8}} & {{60.36}} & 0.554 & {{21.24}} & {{20.90}} & 0/0/100/0 \\
always-Q4 & 100.0 & {{20.36}} & 0.222 & {{39.67}} & {{39.67}} & 0/0/0/100 \\
random & 100.0 & {{42.46}} & 0.421 & {{17.57}} & {{17.57}} & {{3/20/21/56}} \\
{{most exp.}} & {{100.0}} & {{\textbf{58.88}}} & {{0.558}} & {{\textbf{1.15}}} & {{\textbf{1.15}}} & {{13/52/0/35}} \\
\textit{ours} & 100.0 & {{53.30}} & 0.524 & {{6.73}} & {{6.73}} & {{7/32/26/35}} \\
\hline
\multicolumn{7}{|p{230pt}|}{\textbf{C = 4.0 GBitOps} — {{14,998}} feasible architectures} \\
\hline
Oracle & 100.0 & {{85.98}} & {{1.816}} & 0.00 & {{0.00}} & 55/17/27/0 \\
always-Q1 & {{64.8}} & {{83.89}} & {{2.214}} & 0.32 & {{31.61}} & 100/0/0/0 \\
always-Q2 & 100.0 & {{81.66}} & 0.890 & 4.33 & {{4.33}} & 0/100/0/0 \\
always-Q3 & 100.0 & {{66.52}} & 0.890 & {{19.46}} & {{19.46}} & 0/0/100/0 \\
always-Q4 & 100.0 & {{20.36}} & 0.222 & {{65.62}} & {{65.62}} & 0/0/0/100 \\
random & 100.0 & {{61.39}} & {{0.958}} & 24.60 & {{24.60}} & {{16/28/28/28}} \\
{{most exp.}} & {{100.0}} & {{\textbf{85.26}}} & {{1.966}} & {{\textbf{0.72}}} & {{\textbf{0.72}}} & {{65/35/0/0}} \\
\textit{ours} & 100.0 & {{77.31}} & {{1.212}} & {{8.67}} & {{8.67}} & 22/36/42/0 \\
\hline
\end{tabular}
\label{tab:scheme_policy}
\end{table}

\subsection{Hardware Mapping}
\begin{figure}
\centerline{\includegraphics[width=3.1in]{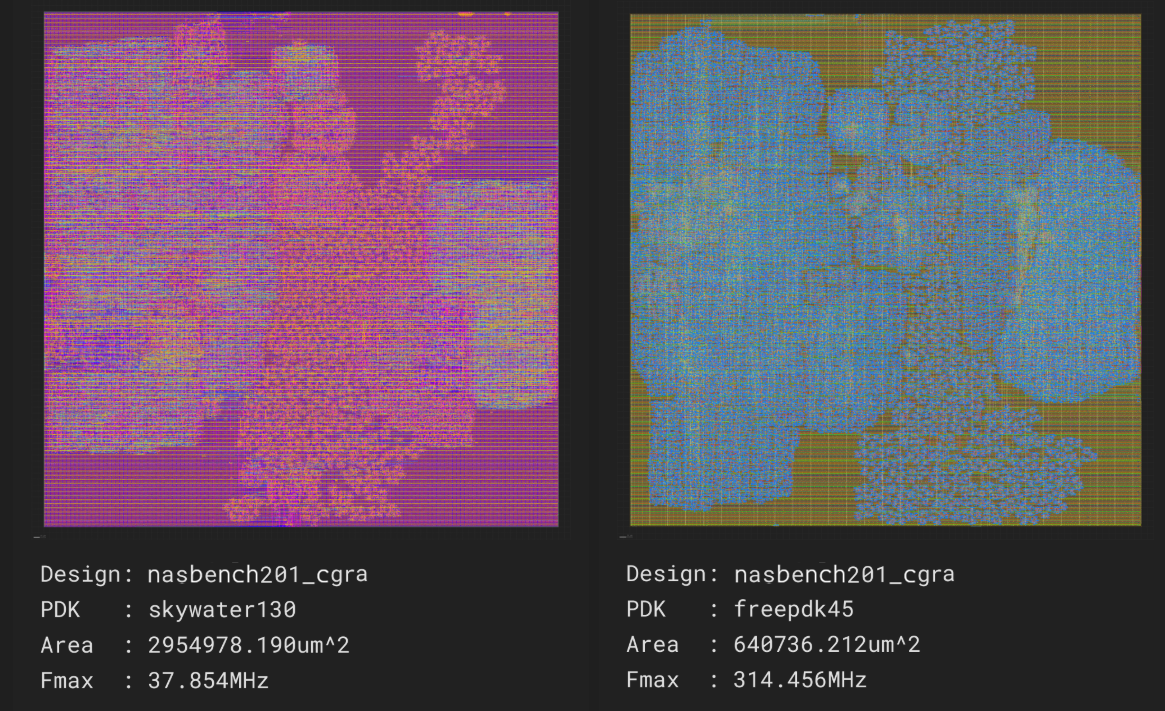}}
\caption{Open-source physical implementation of a CGRA-based NAS-Bench-201 model accelerator on two silicon technologies.}\label{asic}
\end{figure}

Lastly, we present PTQ-CoNAQ's explored FPGA mappings on AMD-Xilinx's ZCU104 board for the Q2 version of \textit{model\_id 6854}, an elected Pareto front architecture, in Table \ref{tab:cgra4ml_vs_finn_int4}{{{{, produced sequentially after PTQ-CoNAQ's BitOps-constrained search.}}}} Both mappings achieve reasonable hardware budget with acceptable performance at 250 MHz. FINN's DSE was initiated with a 500-FPS target, while CGRA4ML’s DSE resulted in a 16×66 CGRA structure with overall utilization of 41.4\%{ \cite{mylonas-ieeesocc2026}}. In addition, Figure \ref{asic} presents two ASIC implementations of the CGRA mapping in two different silicon technologies with the use of the open-source SiliconCompiler tool \cite{siliconcompiler}. {{{The design occupies $2.95$~mm$^2$ at $37.9$~MHz in SkyWater 130~nm and $0.64$~mm$^2$ at $314.5$~MHz in FreePDK45.}}}

\begin{table}
\caption{{Estimated} hardware performance metrics {for} the AMD-Xilinx ZCU104 board.}
\label{table}
\setlength{\tabcolsep}{3pt}
\begin{tabular}{|p{42pt}|p{90pt}|p{90pt}|}
\hline
\textbf{Resource} & \textbf{CGRA4ML design$^{\mathrm{*}}$} & \textbf{FINN design$^{\mathrm{*}}$} \\
\hline
\textbf{LUT} & 57,600/230,400 (25.00\%) & 31,575/230,400 (13.70\%) \\
\textbf{FF} & 92,160/460,800 (20.00\%) & 25,600/460,800 (5.56\%)\\
\textbf{BRAM} & 47/312 (15.06\%)  & 44/312 (14.10\%) \\
\textbf{URAM} & 1/96 (1.04\%) & 0/96 (0.00\%) \\
\textbf{DSP} & 18/1,728 (1.04\%) & 82/1,728 (4.75\%) \\
\hline
\textbf{Throughput} &  3,334 FPS &  565.14 FPS \\
\hline
\multicolumn{3}{p{222pt}}{$^{\mathrm{*}}$Reference model\_id 6854, Q2 PTQ scheme, clocked at 250 MHz.}
\end{tabular}
\label{tab:cgra4ml_vs_finn_int4}
\end{table}

{\section{DISCUSSION} \label{sec:discussion}
{{{The benchmark shows that activation precision governs uniform PTQ, that robustness depends on architecture in ways FP32 accuracy does not reveal, and}}} that no fixed scheme is adequate across {{{budgets. QNP inherits the form of the HAWQ-V2 perturbation metric \cite{hawqv2}; under uniform schemes the bitwidth only rescales such metrics, so they cannot discriminate between schemes, whereas HAWQ-V2, EMQ \cite{emq} and InfoQ \cite{infoq} compare bitwidths across layers of one network. On PTQ-NAS-Bench, FP32 accuracy remains the strongest signal, and scheme selection needs scheme-comparable quantities, such as predicted accuracy, rather than min--max normalised ranks.}}}

{{{\textit{Limitations.} Four limitations bound the conclusions of this work. First, the four released schemes were chosen on CIFAR-10 test accuracy of a single architecture, without a separate validation split. Second, the policy evaluation of Table~\ref{tab:scheme_policy} is in-sample: it includes the predictor-training architectures, and no separate final-evaluation population exists. Third, every result uses the same two-epoch recovery; longer recovery, which changes the Q4 results, is evaluated only on a 150-architecture Q4 sample. Fourth, hardware mapping is demonstrated for one architecture under one scheme, so the hardware results establish feasibility rather than efficiency. We also did not retrain a predictor without the proxy; the ablation of Section~\ref{experiments} compares the training targets directly.}}}}

\section{CONCLUSION} \label{conclusion}
This paper presents PTQ-NAS-Bench, a multi-scheme PTQ extension of NAS-Bench-201, and PTQ-CoNAQ, a {{{{reference pipeline for BitOps-constrained architecture and scheme search with downstream hardware mapping},}}} based solely on PTQ, existing FP32 data and a quantization error estimation proxy. Our work demonstrates the effects of PTQ on the Pareto space analytically{{{{. The returned top-10 candidates lie, at the median, within the best 17--33\% of architectures. However, the benchmark also shows that FP32 accuracy and BitOps, combined with a cost-ordered selection rule, outperform the proxy-driven pipeline, which sets a concrete baseline for future quantization proxies.}}}} Future work will include the extension in other well-known NAS benchmarks and supernets, the enhancement of PTQ-CoNAQ's first stage and the study of more robust PTQ zero-cost proxies.

\begin{IEEEbiography}[{\includegraphics[width=1in,height=1.25in,clip,keepaspectratio]{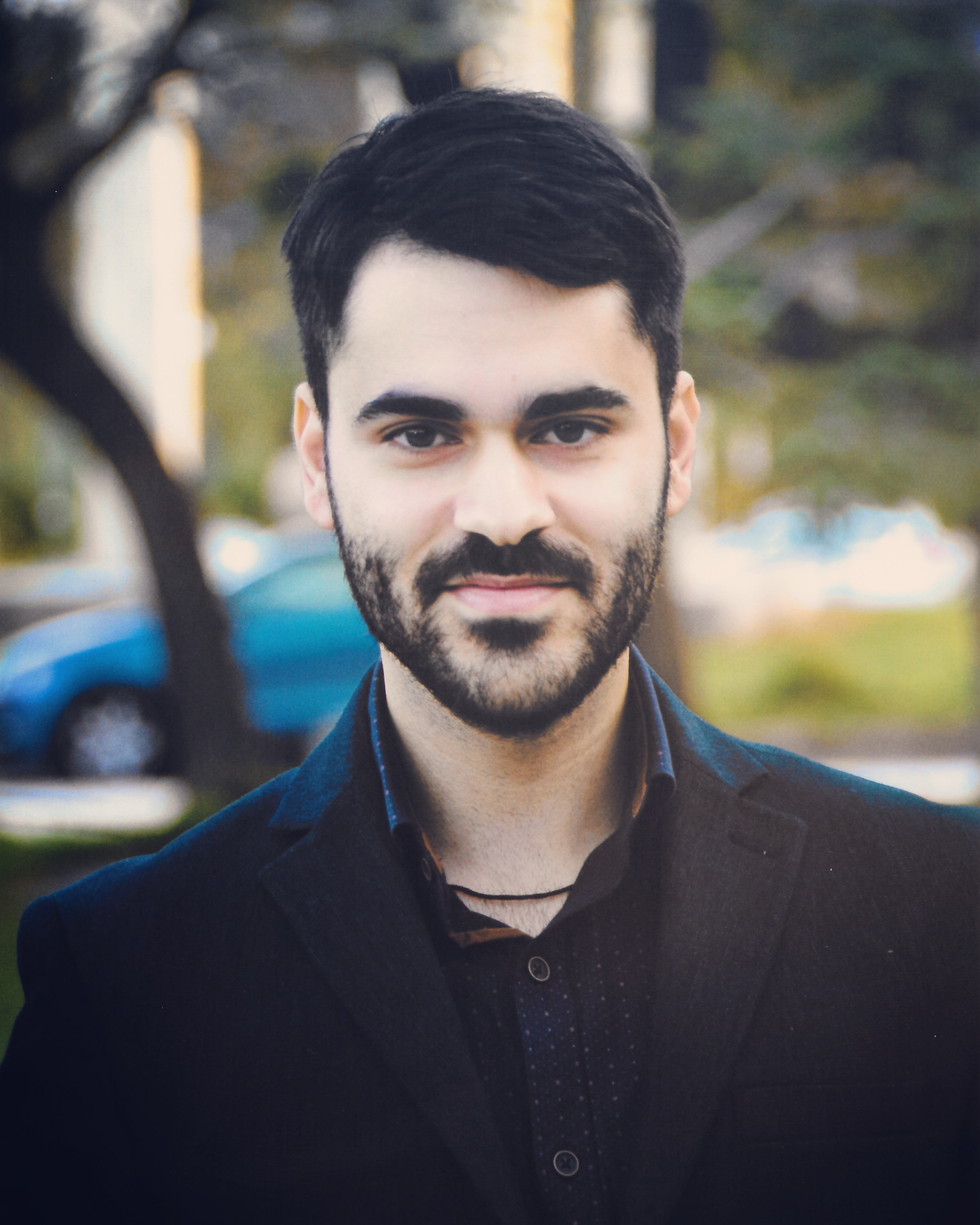}}]
{ELEFTHERIOS MYLONAS}~received the Diploma degree from the Electrical and Computer Engineering Department of University of Patras, in 2019. He is currently pursuing the Ph.D. degree focused on edge intelligence and efficient hardware-accelerated edge AI. He holds research experience in the areas of 5G communications, smart grids, Industry 4.0, AI hardware acceleration. In the past, he served as a Research Engineer at the Department of Research Technology and Development at IPTO S.A. His research interests include Edge AI, efficient cloud-edge architectures, Hardware acceleration, AutoML, Smart grids, Cyber-Physical Systems.
\end{IEEEbiography}

\begin{IEEEbiography}[{\includegraphics[width=1in,height=1.25in,clip,keepaspectratio]{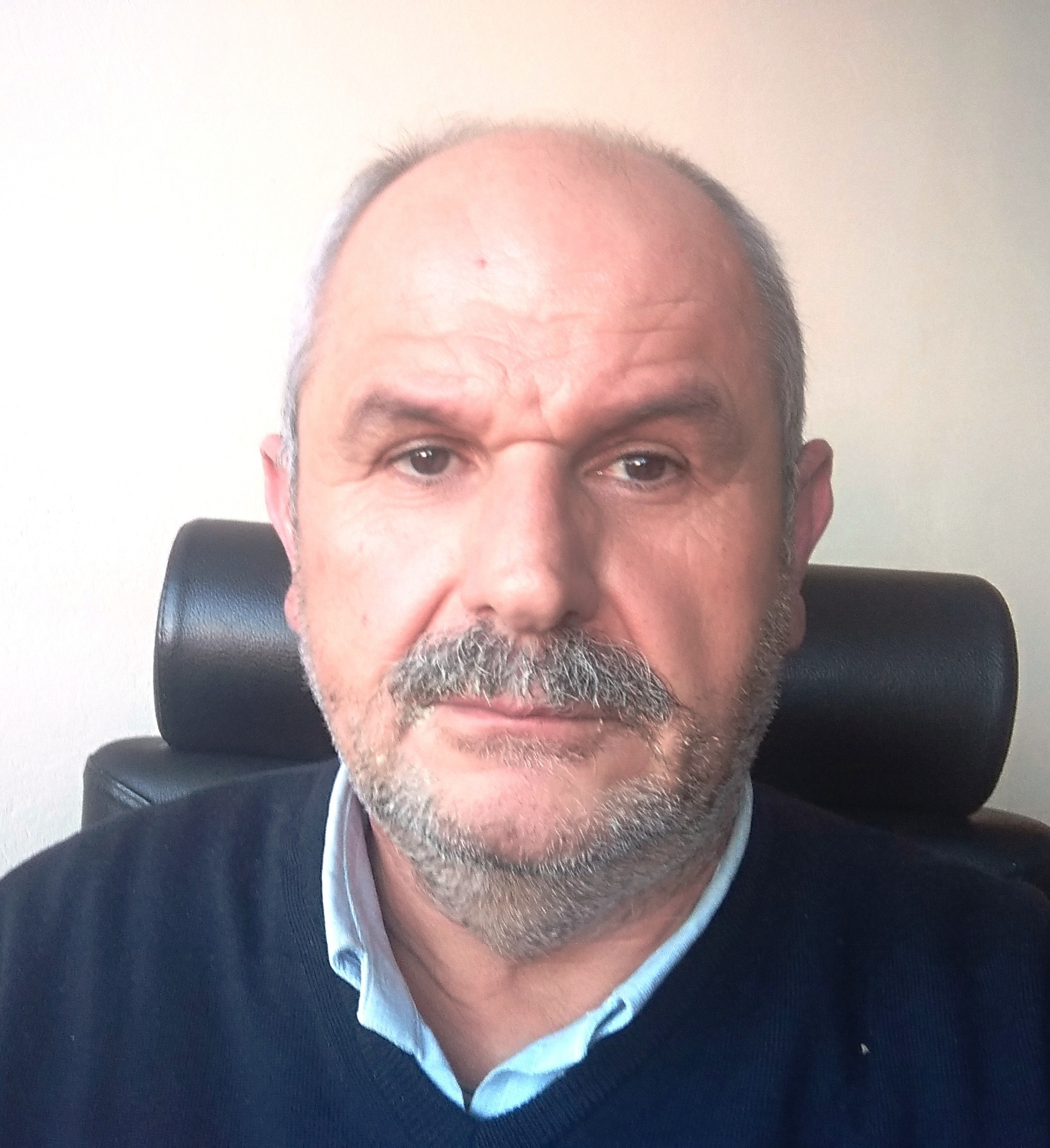}}]
{ALEXIOS BIRBAS}~(Member, IEEE)~received the M.S.E.E. and Ph.D. degrees from the University of Minnesota, Minneapolis, Minnesota, in 1986 and 1988, respectively. He is currently a Professor of electronics with the Department of Electrical and Computer Engineering, and the Director of the Applied Electronics Lab, University of Patras, Patras, Greece. He has also held faculty positions at an assistant professor level with the University of Minnesota and with INPG, Grenoble, France. He has also co-founded two microelectronics related spin-off companies and has supervised 20 Ph.D. Theses. He has also served as a consultant to the Industry and has held industrial managerial positions. He has published more than 150 papers in scientific journals and conference proceedings. His research interests include: device electronics, optoelectronics, noise and fluctuation problems in electronics, RF and mixed signal high speed circuit design and implementation of sensor read out circuits, smart grid electronics and transactive energy systems, smart metering, 5G enabled IoT and Industry 4.0 real time applications, edge computing accelerators, and cyber physical systems.
\end{IEEEbiography}

\begin{IEEEbiography}[{\includegraphics[width=1in,height=1.25in,clip,keepaspectratio]{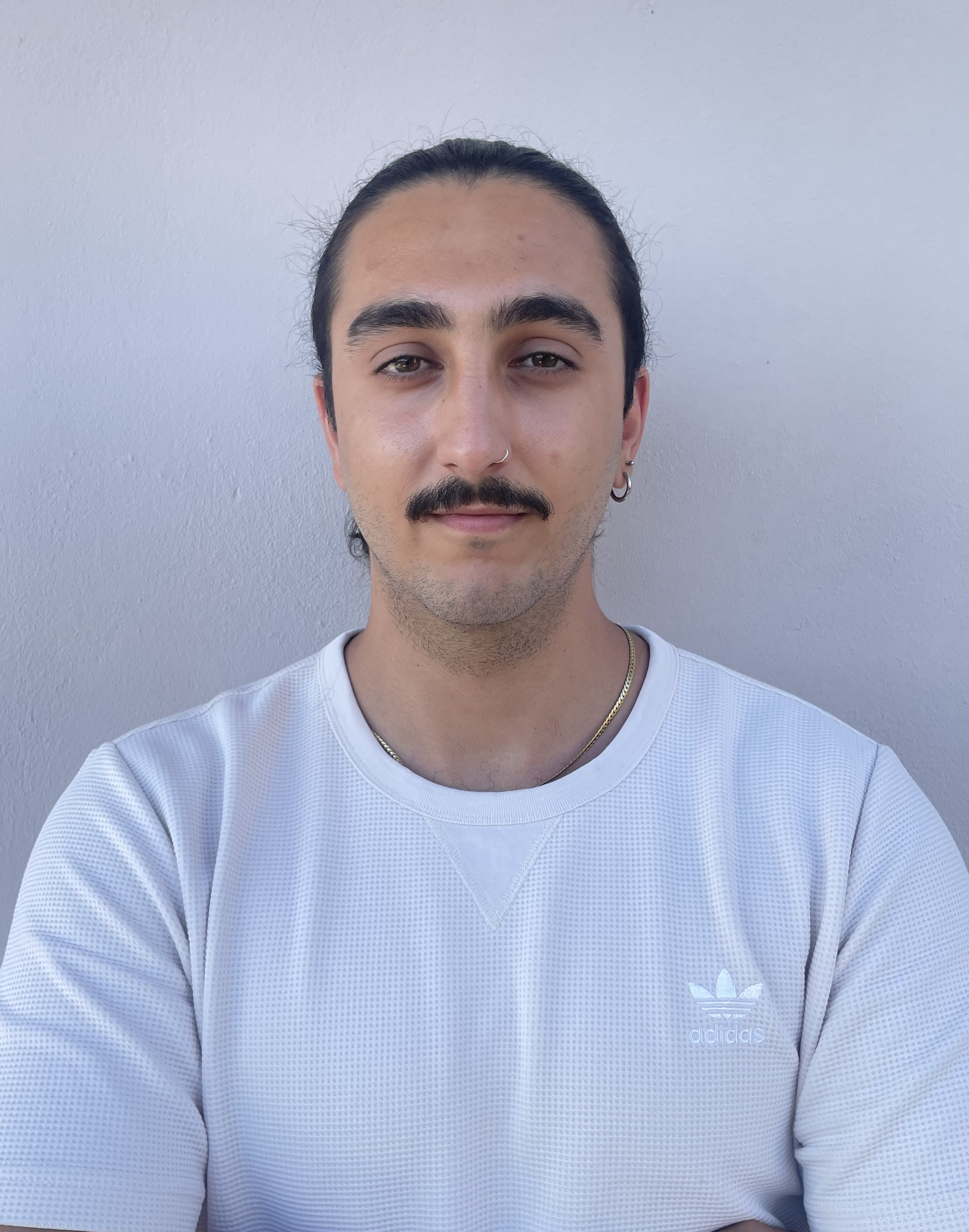}}]
{ANGELOS KOUPRIZAS}~received the Diploma degree from the Electrical and Computer Engineering Department of University of Patras, in 2026. His research interests include Edge AI, NAS, AutoML.
\end{IEEEbiography}

\vfill\pagebreak

\end{document}